\documentclass[10pt,twocolumn,letterpaper,pagebackref,breaklinks,colorlinks,allcolors=cvprblue]{article}

\usepackage{cvpr}              
\usepackage{amsmath,amssymb,amsfonts}
\usepackage{algorithmic}
\usepackage{graphicx}
\usepackage{textcomp}
\usepackage{xcolor}
\usepackage{orcidlink}
\usepackage{longtable,array}
\usepackage{tabularx}
\usepackage{multirow}
\usepackage{makecell}

\definecolor{oursbg}{HTML}{E6F3FF}
\definecolor{arrowcol1}{HTML}{1F77B4}
\usepackage[table]{xcolor}
\usepackage{array}
\usepackage{multirow}
\definecolor{arrowcol}{RGB}{255,76,39}    

\newcommand{\uparr}{\textcolor{arrowcol1}{\ensuremath{\uparrow}}}
\newcommand{\downarr}{\textcolor{arrowcol}{\ensuremath{\downarrow}}}
\usepackage{pifont} 
\newcommand{\xmark}{\ding{55}} 

\definecolor{cvprblue}{rgb}{0.21,0.49,0.74}

\hypersetup{
    pagebackref=true,
    breaklinks=true,
    colorlinks=true,
    allcolors=cvprblue
}

\def\paperID{53} 
\def\confName{CVPR}
\def\confYear{2026}

\title{WildFireVQA: A Large-Scale Radiometric Thermal VQA Benchmark for Aerial Wildfire Monitoring
}
\author{
Mobin Habibpour\,\orcidlink{0009-0001-9271-630X}\qquad
Niloufar Alipour Talemi\,\orcidlink{0009-0000-6881-3671}\qquad
John Spodnik \\
Camren J. Khoury\qquad
Fatemeh Afghah\,\orcidlink{0000-0002-2315-1173}\\
Clemson University\\
{\tt\small \{mhabibp, nalipou, jspodni, camrenk, fafghah\}@clemson.edu}
}

\begin{document}
\maketitle
\begin{abstract}
Wildfire monitoring requires timely, actionable situational awareness from airborne platforms, yet existing aerial visual question answering (VQA) benchmarks do not evaluate wildfire-specific multimodal reasoning grounded in thermal measurements. We introduce WildFireVQA, a large-scale VQA benchmark for aerial wildfire monitoring that integrates RGB imagery with radiometric thermal data. WildFireVQA contains 6,097 RGB–thermal samples, where each sample includes an RGB image, a color-mapped thermal visualization, and a radiometric thermal TIFF, and is paired with 34 questions, yielding a total of 207,298 multiple-choice questions spanning presence and detection, classification, distribution and segmentation, localization and direction, cross-modal reasoning, and flight planning for operational wildfire intelligence. To improve annotation reliability, we combine multimodal large language model (MLLM)-based answer generation with sensor-driven deterministic labeling, manual verification, and intra-frame and inter-frame consistency checks. We further establish a comprehensive evaluation protocol for representative MLLMs under RGB, Thermal, and retrieval-augmented settings using radiometric thermal statistics. Experiments show that across task categories, RGB remains the strongest modality for current models, while retrieved thermal context yields gains for stronger MLLMs, highlighting both the value of temperature-grounded reasoning and the limitations of existing MLLMs in safety-critical wildfire scenarios. The dataset and benchmark code are open-source at \url{https://github.com/mobiiin/WildFire_VQA}.
\end{abstract}

\section{Introduction}
\label{sec:intro}
Wildfires are fast evolving natural disasters that can cause widespread damage to lives, critical infrastructure, and local economies. Effective response depends on timely, actionable situational awareness, where visual understanding from airborne platforms plays a central role. Unmanned aerial vehicles (UAVs) can capture rich observations from active fire scenes, but converting them into reliable decisions remains challenging at scale \cite{afghah2019wildfire}. Recent wildfire-response systems increasingly combine multimodal sensing, simulation, and interactive analytics to support tactical decision making, as in digital-twin frameworks such as FIRETWIN \cite{FIRETWIN}.

Vision language models \cite{carolan2024review, yin2024survey, zhang2024mm, talemi2025style, alipour2025disa} offer a promising direction by combining visual perception with natural language understanding. Among multimodal tasks, visual question answering (VQA) is especially demanding because it requires scene-level comprehension and fine-grained reasoning conditioned on the question \cite{antol2015vqa,marino2019ok}. Aerial VQA is also relatively new, and although recent benchmarks target environmental monitoring, there is still no dataset designed specifically for wildfire monitoring. This gap makes it difficult to measure how well multimodal large language models (MLLMs) handle wildfire-specific reasoning beyond basic fire presence detection. The need is further highlighted by recent efforts on wildfire-aware vision-language reasoning in physics-grounded digital twins for autonomous UAV fire tracking, such as FIRE-VLM \cite{Webb_2026_WACV}.

\begin{figure*}[t]
    \centering
    \includegraphics[width=0.98\linewidth]{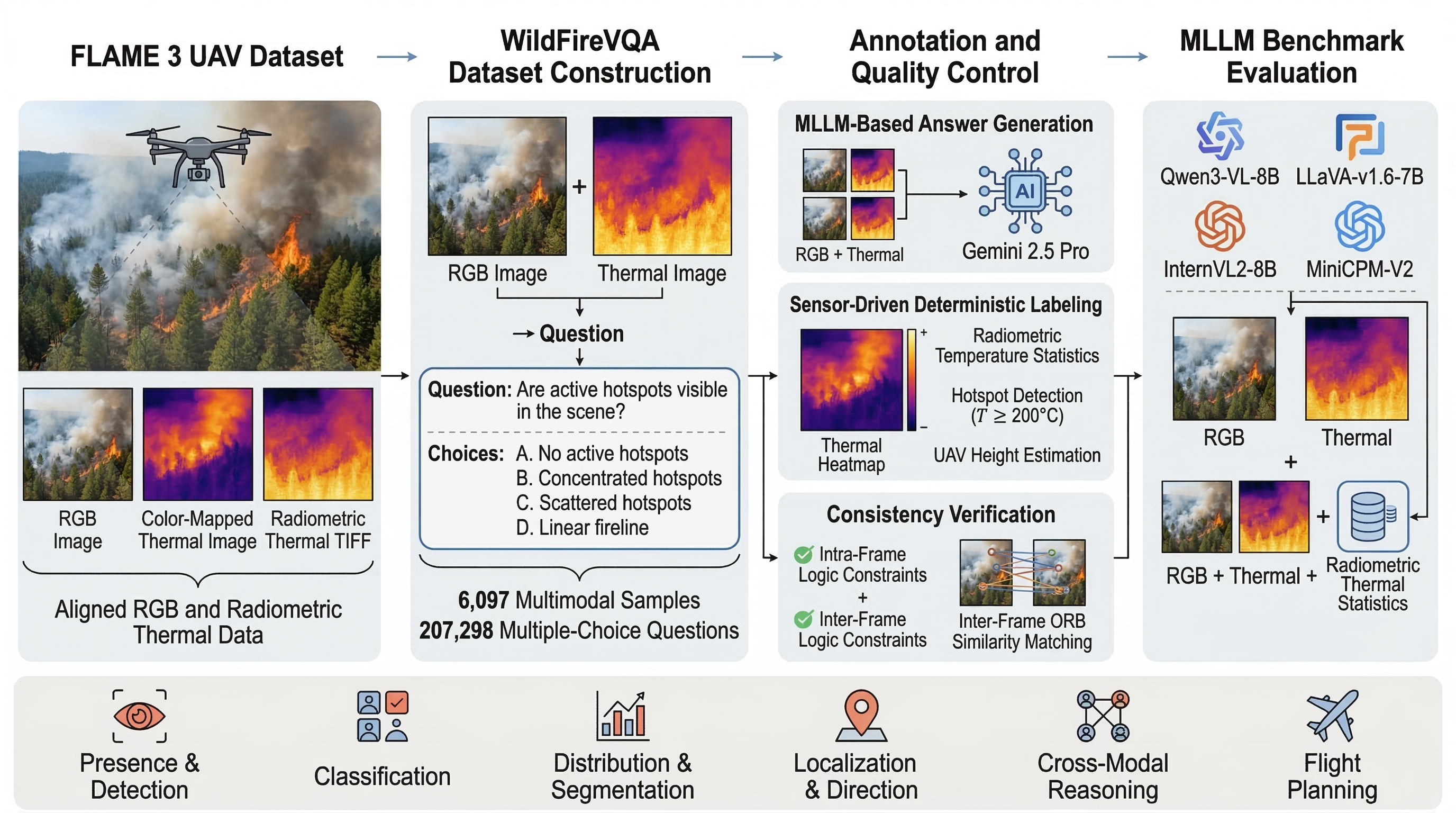}
    \caption{Overview of the WildFireVQA for the operational wildfire intelligence. Unlike standard aerial VQA datasets, we pair standard RGB imagery (left) with radiometric thermal data (right) to enable temperature-grounded reasoning. We provide 34 operationally relevant questions per frame, spanning basic detection to complex flight planning.}
    \label{fig:Overview}\vspace{-1mm}
\end{figure*}

To address this need, we introduce WildFireVQA, a large scale VQA benchmark for UAV-based wildfire monitoring that integrates RGB imagery with radiometric thermal data. Unlike benchmarks that rely only on visual appearance \cite{zheng2021mutual, li2024hrvqa}, WildFireVQA supports temperature-grounded reasoning by providing radiometric thermal TIFFs with per-pixel temperature values in addition to color-mapped thermal visualizations aligned with RGB frames. The benchmark contains 6,097 RGB--thermal samples, with 34 questions per sample, resulting in 207,298 multiple-choice questions spanning detection, classification, segmentation-related queries, localization, cross-modal reasoning between RGB and thermal cues, and flight-planning questions (Fig.~\ref{fig:Overview}).

Constructing a reliable benchmark for a safety-critical domain requires careful label validation. WildFireVQA pairs automated answer proposals with manual review and correction, and further enforces consistency through inter-image and intra-image checks. For inter-image verification, we leverage ORB feature matching \cite{rublee2011orb} to identify overlapping frames and require that answers to identical questions remain consistent across matched views. For intra-image verification, we ensure that answers across different questions for the same frame do not contradict each other.

Alongside the dataset, we provide an evaluation protocol for representative MLLMs under multiple inference settings. Our main-paper experiments focus on zero-shot evaluation across input modalities and retrieval augmentation settings; additional in-context learning \cite{dong2024survey} analyses are provided in the supplementary material. These experiments establish baselines, highlight failure modes, and show how modality choice and retrieved thermal context affect model behavior in a high-stakes environment. By grounding VQA in multimodal wildfire imagery and operationally meaningful questions, WildFireVQA complements existing aerial VQA benchmarks and fills an important gap in current evaluation practice. Our main contributions are as follows:

\begin{itemize}
\item We propose WildFireVQA, a large scale VQA benchmark built on paired RGB imagery and radiometric thermal measurements, enabling temperature grounded question answering for wildfire monitoring.
  \item We design 34 per-frame questions tailored to wildfire monitoring in binary and multiple-choice formats, with carefully curated distractors, and improve annotation reliability through manual verification and inter-frame and intra-frame consistency checks.
  \item We provide a comprehensive evaluation protocol and baselines for representative MLLMs across inference settings, analyzing the effects of modality inputs and in context learning to reveal strengths and failure modes in safety critical wildfire scenarios.
\end{itemize}

\section{Related Work}

\subsection{Remote Sensing VQA Benchmarks}

Remote sensing VQA enables natural-language querying of overhead imagery beyond fixed label sets. Early benchmarks such as RSVQA \cite{lobry2020rsvqa} introduced large-scale image-question-answer triplets generated automatically from geospatial sources, while later datasets such as RSIVQA \cite{zheng2021mutual} and HRVQA \cite{li2024hrvqa} emphasized finer spatial reasoning and higher-resolution aerial imagery. More recent efforts, including VQA-TextRS \cite{al2022open}, RSVLM-QA \cite{zi2025rsvlm}, and VRSBench \cite{li2024vrsbench}, further broaden question diversity and annotation richness for evaluating modern vision-language models. Several remote sensing VQA benchmarks are also motivated by monitoring needs, especially disaster response: FloodNet \cite{rahnemoonfar2021floodnet} provides UAV imagery collected after Hurricane Harvey for post-flood scene understanding and damage assessment, while RescueNet-VQA \cite{sarkar2023rescuenet} targets UAV-based post-disaster damage assessment through structured question answering. Despite this progress, existing remote sensing VQA benchmarks largely focus on generic land-cover or post-disaster assessment, remain predominantly RGB-centric, and, to our knowledge, do not offer a dedicated benchmark for wildfire monitoring that explicitly evaluates thermal reasoning, observability under smoke, and UAV decision-relevant interpretation. FLAME~3 advances wildfire datasets by providing paired RGB and radiometric thermal UAV imagery with per-pixel temperature estimates, but it is not formulated as a VQA benchmark \cite{hopkins2024flame}. These gaps motivate WildFireVQA, a wildfire-specific multimodal VQA benchmark for systematically evaluating fire-monitoring capabilities in FLAME~3 imagery.

\subsection{Wildfire Monitoring Benchmarks}
Public wildfire monitoring datasets span multiple sensing platforms and annotation granularities, reflecting needs from early warning to tactical decision support. For aerial platforms, early benchmarks focused on recognizing fire and smoke cues in cluttered scenes with rapid viewpoint changes, occlusions, and strong illumination variations. The FLAME dataset provides UAV imagery collected during prescribed burns and supports both fire classification and segmentation, including large-scale frame-level Fire/Non-Fire labels and pixel-wise fire masks for fine-grained delineation of flame regions in aerial views \cite{shamsoshoara2021aerial}. FLAME~2 \cite{hopkins2023flame} broadened this setting by releasing paired RGB and thermal UAV imagery at scale and benchmarking learning strategies that exploit multi-spectral inputs for wildfire detection and monitoring beyond purely RGB-based perception.

Beyond the FLAME~1 and 2 series, more recent aerial datasets diversify scene conditions, label types, and mission objectives. UAV image collections have also been introduced for detection and segmentation under varying lighting conditions and with richer class definitions beyond binary fire versus non-fire, enabling more realistic evaluation of false positives in visually similar backgrounds. WIT-UAS targets long-wave infrared detection of crew and vehicle assets in prescribed fire environments and highlights failure modes where thermal fire signatures can induce false positives, motivating fire-aware training data for operational safety monitoring \cite{jong2023wit}. Video-focused resources such as the Boreal Forest Fire dataset further broaden coverage for smoke detection and segmentation by providing annotated UAV images and videos collected at multiple prescribed burning events \cite{pesonen2025boreal}.

FLAME~3 further advances aerial wildfire datasets by introducing synchronized RGB imagery and radiometric thermal data with per-pixel temperature estimates stored in thermal TIFFs, along with thermal visualizations for human inspection. The dataset spans multiple prescribed fire events across diverse fuel and environment types and additionally includes nadir thermal plot recordings for characterizing spatiotemporal fire dynamics. Its radiometric temperature values enable temperature-driven screening and labeling workflows, for example using conservative thresholds to separate clear No-Fire imagery from active-fire imagery before expert review. Compared with prior palette-based thermal datasets, FLAME~3 preserves quantitative thermal information, enabling temperature-aware supervision, more objective validation of fire intensity and hotspots, and learning tasks that require true thermal reasoning rather than color-mapped proxies \cite{hopkins2024flame}.

\section{The WildFireVQA Benchmark}

\subsection{Image Collection}
WildFireVQA is built on FLAME 3 dataset, which provides synchronized aerial visible spectrum (RGB) imagery and radiometric thermal imagery collected by UAVs at rural prescribed burns \cite{hopkins2024flame}. A key advantage of FLAME 3 is that it provides radiometric thermal TIFFs that support per-pixel temperature values, rather than only palette-based thermal visualizations, which enables stronger validation for thermal-driven queries, since hazardous hotspots can remain detectable in temperature even when flames are visually weak or partially obscured, for example when dense smoke masks visible combustion while the TIFF still indicates concentrated regions exceeding high temperature thresholds. 
To support both human readability and algorithmic use, radiometric thermal measurements are stored as single band thermal TIFF rasters, and a corresponding thermal JPEG is produced by mapping temperature values through a color map. In WildFireVQA, we use paired RGB and IR imagery from three prescribed burns: Sycan Marsh, Willamette, and Shoetank. For each frame, we provide three aligned modalities: an RGB image, a color mapped thermal visualization, and a radiometric thermal TIFF that enables temperature grounded queries. These pixel wise temperature measurements provide a stronger basis for validating fire related queries, because radiometric thermal data can reveal elevated heat signatures from smoldering fuel or obscured hotspots even when visible flames are not clearly discernible in RGB imagery.


\subsection{Question Generation}
We design the WildFireVQA question set to benchmark UAV-based wildfire understanding with both binary and multiple-choice items, each paired with carefully constructed distractor options. To reduce positional and language-model biases, we randomly permute answer choices per question instance. We adopt the multiple-choice format following recent MLLM evaluation practice, which reports that constrained outputs offer a more standardized and reliable benchmarking protocol than unconstrained free-form generation \cite{li2024seed, liu2024mmbench}.

Assessing MLLMs as assistants for operational wildfire intelligence requires capabilities beyond detecting fire presence. In practice, analysts must identify fire and smoke cues, interpret fire behavior and fuels, localize hazards, estimate affected regions, and reason about flight safety under limited observability. To capture these requirements, we generate an initial pool of candidate questions by prompting multiple MLLMs with representative bimodal samples from FLAME 3, comprising RGB imagery and color-mapped thermal images. Using targeted prompt engineering, we elicit questions that prioritize actionable, scene-grounded wildfire interpretation, resulting in approximately 50 candidates.

\begin{figure*}
    \centering
    \includegraphics[width=1\linewidth]{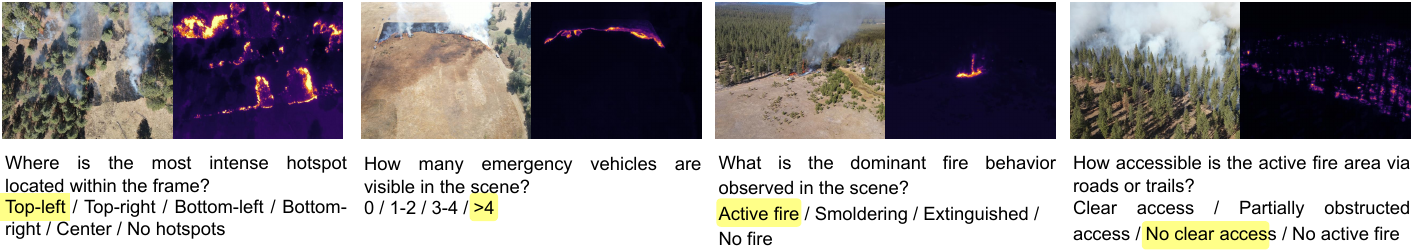}
    \caption{Random examples of RGB-Thermal question-answer quadruplets in WildFireVQA. Each example shows an aligned RGB and radiometric thermal image pair, the associated question, and its ground-truth answer.}
    \label{fig:sample}\vspace{-2mm}
\end{figure*}

We then perform iterative manual curation across diverse samples from all three burns. We remove questions that are unanswerable, unsupported by the imagery, or of limited operational value. For retained items, we refine answer choices to keep distractors informative and non-trivial while avoiding subjective or ambiguous options. We further add expert-designed questions to fill gaps identified during curation. This process yields a final set of 6 binary and 28 multiple-choice questions organized into six task categories aligned with wildfire response needs (see examples in Fig.~\ref{fig:sample}). These categories capture the main forms of reasoning required for operational wildfire intelligence, as summarized below. The complete list of questions is provided in the supplementary material.

\noindent{\textbf{Presence and Detection.}} Determines whether salient wildfire cues and operationally relevant scene elements are present, including fire and smoke indicators, thermal activity, nearby assets and infrastructure, potential suppression resources, and safety-critical context.

\noindent{\textbf{Classification.}} Assigns discrete labels that summarize scene state and environment, including fire behavior, vegetation and canopy characteristics, fuel condition and structure, accessibility, and the types of visible assets (for example, vehicles and built structures).

\noindent{\textbf{Distribution and Segmentation.}} Characterizes how fire, fuels, vegetation, smoke, and heat are distributed across the frame, including pattern-level descriptors and percentage-based coverage estimates (for example, area fractions above specified temperature thresholds).

\noindent{\textbf{Localization and Direction.}} Provides coarse spatial grounding and directional interpretation, including region-level localization of hotspots, vegetation concentrations, smoke origin, and man-made elements, as well as direction cues inferred from  motion patterns such as smoke drift.

\noindent{\textbf{Cross-Modal Reasoning.}} Evaluates joint reasoning over RGB and thermal signals, including cases where one modality reveals information missing in the other, handling occlusion, limited observability, and temperature-informed interpretation of fire activity and heat-bearing fuels.

\noindent{\textbf{Flight Planning.}} Captures UAV operational reasoning, including viewpoint and altitude regimes, environmental and illumination constraints, safety risks near flames or smoke, and scene factors that affect stable and safe flight execution.

\subsection{Answer Generation}
For dataset-wide annotation, we use Gemini 2.5 Pro \cite{comanici2025gemini} to answer all questions for every multimodal sample. Each annotation prompt contains the RGB image, the aligned color-mapped thermal visualization, and a retrieved radiometric temperature summary computed from the paired thermal TIFF. This summary includes the minimum, maximum, and standard deviation of pixel temperatures, together with the percentage of pixels exceeding 200$^\circ$C and 400$^\circ$C. The prompt then presents the question and its candidate answers and instructs the model to select the correct option. By combining visual evidence with compact radiometric statistics, this annotation pipeline preserves physically grounded thermal context while maintaining a standardized multiple-choice labeling protocol.

After a comprehensive investigation of the MLLM-generated labels, we group questions into three categories based on the supervision required for accurate and reliable annotation: (i) questions that remain challenging for current MLLMs but can be answered deterministically from metadata using dedicated mathematical formulations or machine learning pipelines (e.g., Consistency of hotspot intensity, Estimated flight altitude category), which also serve as a useful axis for future evaluation in wildfire monitoring; (ii) questions that are difficult for MLLMs and cannot be inferred from the available metadata via deterministic rules or standard learning methods (e.g., percentage of vegetation affected by fire (burning or burned), moisture level of vegetation), for which we employ additional manual effort and expert verification to audit the labels; and (iii) questions that the MLLM answers correctly in nearly all cases when supported by our prompt design and the paired thermal modality (e.g., dominant fire behavior (Active/Smoldering/Extinguished/No fire), smoke visible (Yes/No)). In the following section, we describe the algorithms used for answer generation in more detail.

\subsection{Sensor-Driven and Deterministic Labeling}

\begin{figure}
    \centering
    \includegraphics[width=1\linewidth]{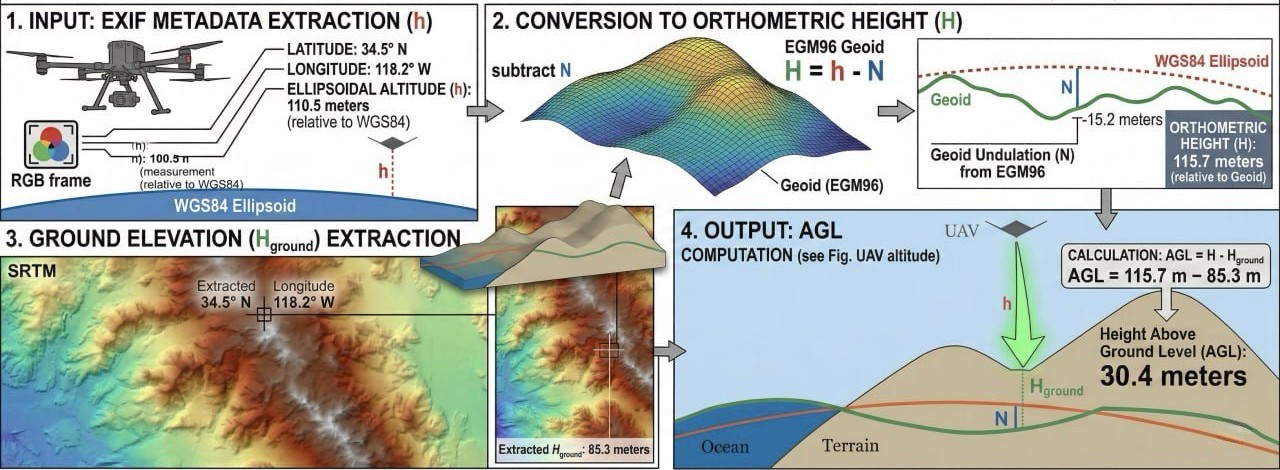}
    \caption{UAV altitude above ground level calculation.}
    \label{fig: UAV altitude}\vspace{-2mm}
\end{figure}

\noindent\textbf{UAV Height Estimation.}
To answer the UAV height above ground level (AGL) question, we use a deterministic sensor derived pipeline rather than visual estimation. For each RGB frame, GPS latitude, longitude, and altitude are extracted from EXIF metadata, where altitude $h$ is reported relative to the WGS84 ellipsoid. We convert ellipsoidal altitude to orthometric height $H$ by subtracting the geoid undulation $N$ from the EGM96 model. The orthometric ground elevation $H_{\text{ground}}$ at the same location is obtained from Shuttle Radar Topography Mission (SRTM) data \cite{farr2007shuttle}. We then compute AGL as the difference between the UAV orthometric altitude and the local ground elevation (see Fig.~\ref{fig: UAV altitude}).

\noindent\textbf{Thermal Hotspot Definition and Detection.}
We detect active wildfire regions from radiometric thermal imagery using a physically grounded, deterministic, altitude-aware formulation that operates directly on per-pixel temperature measurements, avoiding reliance on qualitative visual cues or language-model interpretation. For each frame, we use radiometric thermal TIFF data in which every pixel encodes an absolute temperature (in degrees Celsius). Let $T(x,y)$ denote the temperature at pixel location $(x,y)$, enabling thresholding in temperature space without dependence on color palettes or relative intensity normalization. A pixel is classified as thermally active if its temperature exceeds a fixed physical threshold:
\begin{equation}
\label{eq:hot_pixel}
\mathcal{H}(x,y) =
\begin{cases}
1, & T(x,y) \ge 200^\circ\mathrm{C}, \\
0, & \text{otherwise}.
\end{cases}\vspace{-1mm}
\end{equation}
The $200^\circ$C threshold effectively captures actively burning regions such as flame fronts \cite{hopkins2024flame}, torching vegetation, and crown fire activity while suppressing warm background terrain and non-fire heat sources. We then aggregate thermally active pixels into candidate hotspots via 8-connected component analysis. Each connected component $\mathcal{C}_i$ corresponds to a contiguous thermally active region consisting of $N_i$ pixels:
\begin{equation}
\label{C}
\mathcal{C}_i = \{(x,y) \mid \mathcal{H}(x,y)=1 \}.
\end{equation}

To ensure robustness across varying UAV altitudes, we convert pixel counts to ground-projected physical dimensions using the UAV above-ground altitude $H$ and the thermal camera field of view. With a field-of-view based approximation, the ground sampling distance is:
\begin{equation}
\label{eq:gsd}
\mathrm{GSD} = \frac{2H \tan(\theta/2)}{W},\vspace{-1mm}
\end{equation}
where $\theta$ is the thermal lens diagonal field of view (61$^\circ$ for the DJI~M30T) and $W$ is the thermal image width in pixels. The physical ground area of a candidate hotspot is computed as:
\begin{equation}
\label{eq:area}
A_i = N_i \cdot \mathrm{GSD}^2,\vspace{-1mm}
\end{equation}
We summarize its spatial extent by the equivalent circular, ground-projected radius, $r_i = \sqrt{A_i/\pi}$. A connected component $\mathcal{C}_i$ is considered a valid thermal hotspot if:
\begin{equation}
r_i \ge r_{\min}.\vspace{-1mm} \quad \text{and} \quad N_i \ge N_{\min}.
\end{equation}
\noindent where $r_{\min}=0.75$\,m, and $N_{\min}=5$ pixels enforcing a small minimum pixel count in practice to suppress isolated sensor artifacts and numerical noise. These constraints reduce spurious detections caused by sensor noise, isolated embers, or single-pixel artifacts while preserving physically meaningful fire activity.

\noindent\textbf{Detection of Isolated Heat Sources.}
Building on the hotspot definition, we analyze the spatial organization of valid thermal hotspots to identify isolated heat sources away from the main fire perimeter.
For each frame, we extract hotspot instances $\{\mathcal{C}_i\}_{i=1}^{N}$ with centroid $\mathbf{c}_i=(x_i,y_i)$ in pixel coordinates and physical area $A_i$.
We cluster hotspots via single-linkage using centroid distances in meters:
\begin{equation}
d_{ij} = \sqrt{ \big((x_i-x_j)\cdot \mathrm{GSD}\big)^2 + \big((y_i-y_j)\cdot \mathrm{GSD}\big)^2 } ,
\end{equation}
where $\mathrm{GSD}$ is the altitude-dependent ground sampling distance (m/pixel).
Two hotspots are merged if $d_{ij}\le d_{\mathrm{merge}}$, with $d_{\mathrm{merge}}=10$\,m, yielding clusters $\{\mathcal{K}_k\}_{k=1}^{K}$. We define the main fire cluster as the cluster with the largest total ground-projected area:\vspace{-1mm}
\begin{equation}
k^\star = \arg\max_{k} \sum_{\mathcal{C}_i \in \mathcal{K}_k} A_i, 
\qquad
\mathcal{M} = \mathcal{K}_{k^\star}.\vspace{-1mm}\vspace{-1mm}
\end{equation}
A non-main cluster $K_k$ is treated as an isolated heat source when the minimum centroid distance between any hotspot in $K_k$ and any hotspot in the main cluster $M$ is at least $30\,\text{m}$ (using the same meter-scale centroid distance defined above); if any non-main cluster satisfies this criterion, the frame is labeled as containing isolated heat sources.


\noindent\textbf{Spatial Distribution of Active Hotspots.}
Given the set of thermal hotspot instances extracted in Eq.~\ref{C}, the goal is to characterize the within-frame spatial organization of active burning in a physically grounded, altitude-aware manner. Let $\{\mathcal{C}_i\}_{i=1}^{N}$ denote the detected hotspots, where each hotspot $\mathcal{C}_i$ is associated with a ground-projected centroid $\mathbf{P}_i \in \mathbb{R}^2$ (in meters) and physical area $A_i$, computed using the GSD. If no valid hotspots are detected ($N=0$), the spatial distribution is labeled as \emph{No active hotspots}.

To identify elongated fire structures (e.g., firelines), centroid collinearity is assessed via principal component analysis (PCA) on $\{\mathbf{P}_i\}$. Let $\lambda_1 \ge \lambda_2$ be the eigenvalues of the covariance matrix of the centroid coordinates. Then, a linearity score is defined as $L=\lambda_1/(\lambda_1+\lambda_2)$. For $N \ge 3$, the distribution is classified as \emph{Linear} if $L \ge \tau_{\mathrm{lin}}$ and the maximum pairwise centroid distance exceeds a minimum extent threshold $d_{\mathrm{lin}}$. For the special case $N=2$, the distribution is labeled as \emph{Linear} if the inter-hotspot distance exceeds $d_{\mathrm{lin}}$. In all experiments, $\tau_{\mathrm{lin}}=0.90$ and $d_{\mathrm{lin}}=20$\,m.

For configurations that do not satisfy the linearity criteria, compactness is determined by comparing the spatial extent of hotspots to the equivalent radius of their combined physical area. Let $D_{\max}$ denote the maximum pairwise distance between hotspot centroids. The combined hotspot area and its equivalent radius are defined as:
\begin{equation}
A_{\mathrm{tot}} = \sum_{i=1}^{N} A_i, \qquad
r_{\mathrm{eq}} = \sqrt{\frac{A_{\mathrm{tot}}}{\pi}}.\vspace{-1mm}
\end{equation}
The distribution is classified as \emph{Concentrated} if
\begin{equation}
D_{\max} \le \alpha \, r_{\mathrm{eq}},
\end{equation}
and as \emph{Scattered} otherwise, with $\alpha=4.0$ used throughout. Therefore, the spatial distribution label (SDL) is assigned by the following rule:\vspace{-1mm}
\begin{equation}
\text{SDL} =
\begin{cases}
\text{No active hotspots}, & N = 0, \\
\text{Linear}, & \text{linearity criteria satisfied}, \\
\text{Concentrated}, & D_{\max} \le \alpha \, r_{\mathrm{eq}}, \\
\text{Scattered}, & \text{otherwise}.
\end{cases}
\end{equation}

\noindent\textbf{Hotspot Intensity Consistency.}
Given thermal hotspots extracted as in Eq.~\ref{C}, hotspot intensity consistency assesses whether hotspots in a scene exhibit comparable thermal intensity or contain clearly distinct intensity levels. For each hotspot $\mathcal{C}_i$, a representative intensity is defined as
\begin{equation}
I_i = \max_{(x,y)\in \mathcal{C}_i} T(x,y),
\end{equation}
where $T(x,y)$ is the temperature (in $^\circ$C) at pixel $(x,y)$. Using the maximum temperature provides a physically meaningful indicator of local fire intensity while remaining invariant to hotspot size. For a scene with $N$ valid hotspots, intensity variation is measured using a robust coefficient of variation based on the median absolute deviation (MAD). Let $\{I_i\}_{i=1}^{N}$ denote hotspot intensities, with median $\tilde{I}=\mathrm{median}(\{I_i\})$ and $\mathrm{MAD}=\mathrm{median}(\{|I_i-\tilde{I}|\})$. The robust coefficient of variation is defined as:

\begin{equation}
\mathrm{rCV} = \frac{1.4826 \cdot \mathrm{MAD}}{\max(\tilde{I}, \epsilon)},
\end{equation}
where $\epsilon$ is a small constant for numerical stability. The hotspot intensity consistency label (HICL) is assigned as:
\begin{equation}
y_{\mathrm{int}}=
\begin{cases}
\text{\emph{No active hotspots}}, & N=0,\\
\text{\emph{Similar intensity}}, &
\begin{aligned}[t]
N\ge 1 \ \land\ (&\mathrm{rCV}\le\tau_{\mathrm{sim}} \\
&\lor\ \Delta I\le\Delta T_{\mathrm{sim}}),
\end{aligned}\\
\text{\emph{Clearly different}}, & \text{otherwise},
\end{cases}
\end{equation}
\noindent\text{where }$\Delta I=\max_i I_i-\min_i I_i$, $\tau_{\mathrm{sim}} = 0.10$ and $\Delta T_{\mathrm{sim}} = 20^\circ\mathrm{C}$. Beyond hotspot-level comparisons, scene-level extreme-temperature coverage is computed from the radiometric thermal TIFF as the fraction of pixels exceeding fixed thresholds. Let $N_{\mathrm{pix}}$ denote the total number of pixels. The percentage of the scene above a threshold $\tau$ is computed as:
\begin{equation}
p_{\tau} = 100 \cdot \frac{1}{N_{\mathrm{pix}}}\sum_{x,y}\mathbb{I}\left[T(x,y)\ge \tau\right],
\end{equation}
where $\mathbb{I}[\cdot]$ is the indicator function. WildFireVQA includes two coverage queries: one reports $p_{400}$ binned into $\{\text{None}, <2\%, 2\text{--}4\%, 4\text{--}6\%, >6\%\}$, and the other reports $p_{200}$ binned into $\{\text{None}, <5\%, 5\text{--}10\%, 10\text{--}15\%, >15\%\}$.

\noindent\textbf{Location of the Most Intense Hotspot.} Using the radiometric thermal map $T(x,y)$ (in $^\circ$C), we define active hotspot pixels as those satisfying $T(x,y)\ge 200^\circ\mathrm{C}$. If no such pixels exist, the answer is \emph{No hotspots}. Otherwise, we identify the hottest hotspot location as the argmax pixel $(x^\star,y^\star)=\arg\max_{x,y} T(x,y)$ restricted to the hotspot mask. We then assign the location category by partitioning the image into five regions: \emph{Center} corresponds to the middle third of the frame in both axes ($x\in[W/3,2W/3)$ and $y\in[H/3,2H/3)$), while the remaining area is split by image midlines into the four quadrants (\emph{Top-left}, \emph{Top-right}, \emph{Bottom-left}, \emph{Bottom-right}).

\noindent\textbf{Emergency Vehicle Counting.}
To obtain reliable ground-truth counts for emergency vehicles (a task where MLLMs can hallucinate or miscount), we use an open-vocabulary object detector instead of language-only inference. Specifically, we apply YOLO-World, a real-time open-vocabulary detector that supports text-prompted category definitions \cite{cheng2024yolo}. Given the RGB frame, we query the detector with prompts related to emergency response vehicles (e.g., \textit{fire truck}, \textit{ambulance}, \textit{police car}, \textit{emergency vehicle}). We aggregate all detections above a confidence threshold and apply class-agnostic non-maximum suppression to merge overlapping boxes across prompts, producing the final count. 

\subsection{Consistency Verification and Quality Control}

After dataset-wide labeling and manual inspection of all annotations, two additional quality-control procedures are applied as a final consistency audit to identify and correct any remaining inconsistencies.

\noindent\textbf{Intra-Frame Cross-Question Consistency Constraints.}
We apply cross-question logic constraints within each frame to detect contradictions among semantically related questions. For example, if the scene is labeled as \emph{No fire}, then hotspot, flame, and fire-base visibility questions should not indicate active burning. Similarly, if \emph{No smoke} is selected for smoke presence, the estimated smoke coverage should be \emph{No smoke}. If \emph{No active hotspots} is predicted, then the hotspot location question must also return \emph{No hotspots}, and hotspot intensity consistency must return \emph{No active hotspots}. Likewise, if no man-made structures are detected, the structure localization question should be \emph{No structures visible}. Together, these rule-based consistency checks, deterministic approaches and manual verification reduce label noise while preserving scalable dataset-wide annotation. 

\noindent\textbf{Inter-Frame Visual Similarity Consistency via ORB Matching.}
We perform an inter-frame, similarity-driven consistency audit to detect label conflicts among visually near-duplicate frames. Since WildFireVQA contains multiple burns, and each burn is organized into \textit{Fire} and \textit{No-Fire} subsets, we treat each subset as a label-consistent seed pool. For every query frame, we extract ORB~\cite{rublee2011orb} keypoints and binary descriptors and match them using a brute-force Hamming-distance matcher. Candidate correspondences are first filtered using Lowe's ratio test (ratio $=0.8$) and then geometrically verified by estimating a homography with RANSAC (reprojection threshold $=20.0$). A pair of frames is considered a near-duplicate if at least $15$ geometrically consistent inlier matches are obtained. ORB is configured to detect up to $8000$ features per image to improve robustness under viewpoint and illumination variations. Matching is performed in a multimodal manner: RGB images are compared first, and thermal images are used as a fallback when RGB matching fails. Frames that satisfy the inlier threshold are grouped as near-duplicates, after which we enforce label agreement within each group by flagging cases where a query frame and any of its near-duplicates are assigned conflicting \textit{Fire} versus \textit{No-Fire} labels, or where fire-relevant answers contradict across the group. This retrieval-based audit complements intra-frame logic by ensuring spatiotemporal stability across the dataset.

\begin{table}[t]
\centering
\caption{Answer choice distribution and random-guess accuracy across WildFireVQA tasks. Random accuracy assumes uniform selection. {\#Q} denotes the number of question types per category.}
\label{tab:choice_random_accuracy}
\setlength{\tabcolsep}{5pt}
\renewcommand{\arraystretch}{0.95}
\resizebox{0.45\textwidth}{!}{%
\begin{tabular}{lccc}
\toprule
\rowcolor{oursbg}
\textbf{Category} & \textbf{\#Q} & \textbf{Avg. Choices} & \textbf{Random Acc. (\%)} \\
\midrule
Presence \& Detection & 8 & 2.38 & 44.79 \\
Classification & 6 & 3.83 & 26.39 \\
Distribution \& Segmentation & 8 & 4.25 & 24.58 \\
Localization \& Direction & 4 & 6.50 & 15.48 \\
Cross-Modal Reasoning & 4 & 4.75 & 21.67 \\
Flight Planning & 4 & 3.50 & 31.25 \\
\midrule
\textbf{Overall} & 34 & 3.97 & 29.03 \\
\bottomrule
\end{tabular}%
}\vspace{-2mm}
\end{table}

\section{Experiments and Analysis}

\subsection{Experimental Setup}
We evaluate four representative MLLMs, namely LLaVA-v1.6-Mistral-7B \cite{liu2023improved}, Qwen3-VL-8B-Instruct \cite{bai2025qwen3}, InternVL2-8B \cite{chen2024internvl}, MiniCPM-V2 \cite{yao2024minicpmvgpt4vlevelmllm}, to establish baselines for WildFireVQA. Because the questions contain different numbers of answer choices, task difficulty varies across categories. We also report the distribution of answer choices and the corresponding random-guess accuracy, computed under uniform selection, in Table~\ref{tab:choice_random_accuracy}. All models are evaluated in a zero-shot setting to assess out-of-the-box generalization on wildfire-specific multimodal reasoning, and performance is measured using accuracy. To systematically evaluate MLLM reasoning on WildFireVQA, we conduct controlled experiments that isolate the effects of input modality, and retrieval augmentation. Retrieval-augmented generation (RAG) appends retrieved radiometric thermal statistics as auxiliary context. This setup distinguishes gains due to visual evidence alone from gains driven by explicit temperature cues in the paired radiometric thermal TIFF, and reveals how models use RGB, thermal structure, and retrieved statistics. \textbf{See supplementary material for full question details, multimodal examples, and additional experiments.}


\begin{table}[t]
\centering
\caption{Comparison of four MLLMs under six input settings on the WildFireVQA wildfire monitoring MCQ benchmark. In the table, \textbf{Qwen3-VL} denotes \textit{Qwen3-VL-8B-Instruct}, \textbf{LLaVA-v1.6} denotes \textit{LLaVA-v1.6-Mistral-7B}, \textbf{InternVL2} denotes \textit{InternVL2-8B}. \textbf{Cls.} = Classification, \textbf{CMR} = Cross-Modal Reasoning, \textbf{D\&S} = Distribution and Segmentation, \textbf{FP} = Flight Planning, \textbf{L\&D} = Localization and Direction, \textbf{P\&D} = Presence and Detection, and \textbf{Overall} = overall accuracy across all tasks.}
\label{tab:all_models_wildfire_results}
\setlength{\tabcolsep}{3pt}
\renewcommand{\arraystretch}{1.02}
\resizebox{0.48\textwidth}{!}{%
\begin{tabular}{llcccccccc}
\toprule
\rowcolor{oursbg}\textbf{Model} & \textbf{Setting} & \textbf{RAG} & \textbf{Overall} & \textbf{Cls.} & \textbf{CMR} & \textbf{D\&S} & \textbf{FP} & \textbf{L\&D} & \textbf{P\&D} \\
\midrule

\multirow{4}{*}{\textbf{Qwen3-VL}}
& RGB & No  & 53.28 & 45.38 & 57.45 & 41.28 & 51.70 & 40.35 & 76.39 \\
& RGB & Yes & 54.76 & 47.67 & 65.77 & 47.20 & 51.07 & 43.47 & 69.64 \\
\cmidrule(lr){2-10}
& Thermal & No  & 33.00 & 31.17 & 28.54 & 27.07 & 22.23 & 23.40 & 52.70 \\
& Thermal & Yes & 38.08 & 30.71 & 49.75 & 39.17 & 29.55 & 16.97 & 51.49 \\
\midrule

\multirow{4}{*}{\textbf{LLaVA-v1.6}}
& RGB & No  & 52.68 & 49.56 & 38.75 & 48.78 & 19.64 & 61.50 & 78.01 \\
& RGB & Yes & 51.88 & 48.55 & 61.60 & 38.78 & 22.68 & 55.26 & 75.51 \\
\cmidrule(lr){2-10}
& Thermal & No  & 38.74 & 31.88 & 31.26 & 20.50 & 16.06 & 51.99 & 70.56 \\
& Thermal & Yes & 41.23 & 42.93 & 28.16 & 29.98 & 21.85 & 46.60 & 64.75 \\

\midrule

\multirow{4}{*}{\textbf{InternVL2}}
& RGB & No  & 45.79 & 43.57 & 31.23 & 42.36 & 24.27 & 38.08 & 72.79 \\
& RGB & Yes & 47.66 & 46.40 & 30.80 & 47.78 & 24.56 & 45.60 & 69.47 \\
\cmidrule(lr){2-10}
& Thermal & No  & 34.21 & 33.89 & 11.87 & 27.71 & 23.34 & 38.63 & 55.34 \\
& Thermal & Yes & 39.21 & 34.92 & 23.78 & 40.84 & 23.84 & 38.38 & 56.60 \\

\midrule

\multirow{4}{*}{\textbf{MiniCPM-V2}}
& RGB & No & 49.51 & 47.26 & 50.41 & 40.31 & 17.63 & 59.77 & 70.73 \\
& RGB & Yes & 45.72 & 46.70 & 36.01 & 34.79 & 21.60 & 48.05 & 71.65 \\
\cmidrule(lr){2-10}
& Thermal & No & 36.10 & 31.65 & 37.33 & 28.04 & 17.12 & 33.34 & 57.76 \\
& Thermal & Yes & 34.86 & 34.75 & 30.45 & 22.32 & 21.43 & 34.11 & 56.77 \\

\bottomrule
\end{tabular}%
}\vspace{-2mm}
\end{table}

\subsection{Experiment Results}
Qwen3-VL-8B achieves the strongest overall performance among all evaluated models, with its best result obtained by RGB with retrieved radiometric thermal statistics at 54.76\% overall accuracy. RGB remains the stronger modality for this model, but gains under both RGB and Thermal indicate that Qwen3-VL can use the retrieved thermal summary as a complementary cue. LLaVA-v1.6-7B performs best with RGB without retrieved radiometric thermal statistics, reaching 52.68\%, suggesting that it relies most effectively on RGB cues alone. Thermal input remains weaker overall for LLaVA, although it becomes more competitive when augmented with retrieved thermal statistics, indicating that RAG helps selectively when the added thermal information is closely aligned with the reasoning required by the question.

InternVL2-8B also shows a generally positive response to retrieval augmentation, with its best performance achieved by RGB with retrieved radiometric thermal statistics at 47.66\%. Although RGB remains its stronger modality, the substantial improvement in the Thermal setting suggests that InternVL2 can use the thermal summary as a complementary signal, with gains concentrated in categories more naturally grounded in thermal evidence. MiniCPM-V2 achieves its best result with RGB without retrieved radiometric thermal statistics at 49.51\%, confirming that RGB is its strongest input modality. Unlike Qwen3-VL and InternVL2, however, MiniCPM-V2 does not benefit from retrieval augmentation, as performance decreases in both RGB and Thermal settings. This suggests that the limitation lies less in the retrieved information itself and more in the model’s weaker ability to combine visual content, question context, answer choices, and structured numeric thermal metadata within a single prompt.


\noindent\textbf{Comparison across models.}
Table~\ref{tab:all_models_wildfire_results} shows that RGB-based settings perform best across all models. Qwen3-VL-8B achieves the highest accuracy at 54.76\%, followed by LLaVA-v1.6-7B (52.68\%), MiniCPM-V2 (49.51\%), and InternVL2-8B (47.66\%). Retrieval augmentation helps Qwen3-VL-8B and InternVL2-8B more consistently, helps LLaVA-v1.6-7B selectively, and degrades MiniCPM-V2, indicating that effective use of retrieved thermal context remains model dependent.

\begin{table}[t]
\centering
\caption{Overall accuracy gain after applying RAG across different MLLMs on the WildFireVQA benchmark. Gains are computed as \textit{Overall (Yes) - Overall (No)} for each input setting. Positive values indicate that RAG improves performance, while negative values indicate performance degradation.}
\label{tab:overall_rag_gain_models}
\setlength{\tabcolsep}{5pt}
\renewcommand{\arraystretch}{1.05}
\resizebox{0.48\textwidth}{!}{%
\begin{tabular}{lcccc}
\toprule
\rowcolor{oursbg}
\textbf{Setting} & \textbf{Qwen3-VL} & \textbf{LLaVA-v1.6} & \textbf{InternVL2} & \textbf{MiniCPM-V2}  \\
\midrule
\textbf{RGB Gain} 
& $+1.48$\uparr 
& $-0.80$\downarr 
& $+1.87$\uparr 
& $-3.79$\downarr 
 \\

\textbf{Thermal Gain} 
& $+5.08$\uparr 
& $+2.49$\uparr 
& $+5.00$\uparr 
& $-1.24$\downarr 
 \\
\bottomrule
\end{tabular}
}\vspace{-2mm}
\end{table}

\noindent\textbf{Effect of RAG and task-level trends.}
Table~\ref{tab:overall_rag_gain_models} shows that the effect of RAG is helpful for stronger MLLMs. In our setting, RAG appends a compact radiometric summary from the paired thermal TIFF, including temperature statistics and the percentage of pixels above thresholds such as 200$^\circ$C and 400$^\circ$C. This provides physically grounded auxiliary context beyond the RGB image and/or color-mapped thermal visualization. Comparing RAG and non-RAG settings therefore separates gains from improved visual reasoning from gains due to explicit sensor-derived temperature cues. Qwen3-VL-8B and InternVL2-8B benefit consistently, especially in the Thermal setting, indicating effective use of structured thermal evidence. The gains are most apparent for questions directly grounded in temperature cues, while less thermal-dependent categories show smaller improvements or slight declines. Notably, the benefit of retrieved thermal metadata depends on the model’s ability to integrate radiometric statistics with multimodal inputs.

\section{Conclusion}

WildFireVQA provides a large-scale radiometric thermal VQA benchmark for UAV-based wildfire monitoring. By combining RGB imagery, color-mapped thermal visualizations, radiometric thermal TIFFs, deterministic sensor-driven labeling, and consistency verification, it enables physically grounded evaluation of wildfire intelligence. Our baselines show that RGB remains the strongest modality overall, while retrieved thermal context benefits stronger MLLMs, highlighting both the value of temperature-grounded reasoning and the limitations of current models in wildfire scenarios. A limitation of the current evaluation protocol is that radiometric thermal statistics are provided explicitly as text under RAG, rather than requiring models to infer them directly from thermal inputs. This design isolates whether current MLLMs can benefit from physically grounded temperature cues once made explicit, while leaving direct end-to-end thermal inference for future work.

\section*{Acknowledgment}
This work was supported in part by NASA award 80NSSC23K1393 and NSF grants CNS-2232048 and CNS-2204445.

{
    \small
    \bibliographystyle{ieeenat_fullname}
    \bibliography{main}
}

\appendix

\maketitlesupplementary

\setlength{\tabcolsep}{2pt}
\renewcommand{\arraystretch}{1.15}

\section*{Supplementary Overview}
This supplementary material complements the main paper by providing a detailed analysis of temperature-grounded retrieval, the complete WildFireVQA question inventory, and additional information on the multimodal inputs and annotation prompt structure used in the evaluation setting.

\begin{table*}[t]
\centering
\caption{Fine-grained comparison of Qwen3-VL \cite{bai2025qwen3} under the RGB setting, with and without retrieved radiometric thermal statistics. We partition the 34 WildFireVQA questions into a temperature-related subset and a remaining-question subset. The temperature-related subset includes questions whose answers depend directly on hotspot presence, fire intensity, or temperature-threshold coverage. Gains are computed as \textit{RAG - No-RAG}.}
\label{tab:qwen_rag_grouped_analysis}
\setlength{\tabcolsep}{5pt}
\renewcommand{\arraystretch}{1.05}
\resizebox{0.95\textwidth}{!}{%
\begin{tabular}{l l c c c c c c}
\toprule
\rowcolor{oursbg}
\textbf{Group} & \textbf{Question IDs} & \textbf{Total} & \textbf{No-RAG Correct} & \textbf{No-RAG Acc.} & \textbf{RAG Correct} & \textbf{RAG Acc.} & \textbf{Gain} \\
\midrule
Temperature-related & PD1, CL1, CMR4, DS3, DS7, DS8 & 36,582 & 22,513 & 61.54 & 26,197 & 71.61 & +10.07\uparr \\
Remaining questions & All others & 170,716 & 87,944 & 51.51 & 87,328 & 51.15 & -0.36\downarr \\
\bottomrule
\end{tabular}%
}
\vspace{5mm}
\end{table*}

\begin{table*}[t]
\centering
\caption{Per-question analysis for for the temperature-related questions (CL1 and CMR4) that show positive gains under retrieved radiometric thermal statistics. Each question contains 6,097 instances. Gains are computed as \textit{RAG - No-RAG}.}
\label{tab:qwen_rag_question_analysis}
\setlength{\tabcolsep}{5pt}
\renewcommand{\arraystretch}{1.05}
\resizebox{0.95\textwidth}{!}{%
\begin{tabular}{c l c c c}
\toprule
\rowcolor{oursbg}
\textbf{ID} & \textbf{Question} & \textbf{No-RAG Acc.} & \textbf{RAG Acc.} & \textbf{Gain} \\
\midrule
CL1 & What is the dominant fire behavior observed in the scene? & 48.93 & 67.69 & +18.76\uparr \\
CMR4 & What is the temperature of the hottest part of the fire in this scene in degrees Celsius? & 58.88 & 88.60 & +29.72\uparr \\
\bottomrule
\end{tabular}%
}
\end{table*}

\section{Fine-Grained Analysis of Temperature-Grounded Retrieval}
To better understand when retrieved radiometric thermal statistics are beneficial, we perform a fine-grained analysis for Qwen3-VL \cite{bai2025qwen3} under the RGB setting by comparing the same model with and without RAG. We partition the 34 WildFireVQA questions into two groups: a temperature-related subset containing PD1, CL1, CMR4, DS3, DS7, and DS8, and a remaining-question subset containing all other questions. The grouped results in Table~\ref{tab:qwen_rag_grouped_analysis} show that the benefit of RAG is highly concentrated on the temperature-related questions. In particular, accuracy on this subset increases from 61.54\% to 71.61\%, yielding a gain of +10.07 points, while the remaining questions show essentially no improvement, changing from 51.51\% to 51.15\%.

This pattern indicates that retrieved radiometric thermal statistics are most useful when the question is directly aligned with the sensor-derived evidence appended to the prompt. In our setting, RAG provides a compact thermal summary from the paired radiometric TIFF, including temperature statistics and the fraction of pixels above physically meaningful thresholds such as 200$^\circ$C and 400$^\circ$C. The per-question results in Table~\ref{tab:qwen_rag_question_analysis} further support this interpretation. In particular, CMR4 benefits substantially because it requires reasoning about the hottest part of the fire, which is strongly supported by explicit temperature cues. CL1 also shows a clear improvement, suggesting that retrieved thermal evidence helps the model better distinguish active fire behavior from weaker or less intense burning conditions. Taken together, these results indicate that the proposed retrieval strategy is most effective when the model must reason about fire intensity and thermally grounded scene state, rather than relying only on visual appearance or broader scene interpretation alone.

\section{Complete Question Set and Answer Space}
\label{sec:mca}

Table~\ref{tab:question_taxonomy_final} lists the complete WildFireVQA question set together with the canonical answer space for each question. The questions are grouped by the six task categories used in the main paper: Presence and Detection, Classification, Distribution and Segmentation, Localization and Direction, Cross-Modal Reasoning, and Flight Planning. Defining a fixed answer space for each question helps reduce ambiguity and supports consistent comparison across models and evaluation settings.

\begin{table*}[t]
\centering
\scriptsize
\setlength{\tabcolsep}{3pt}
\renewcommand{\arraystretch}{1.3}

\caption{Comprehensive list of questions and answer choices grouped by category for wildfire analysis.}
\label{tab:question_taxonomy_final}

\begin{tabularx}{\textwidth}{@{} l X l | l X l @{}}
\toprule
\textbf{ID} & \textbf{Question} & \textbf{Answer Choices} & \textbf{ID} & \textbf{Question} & \textbf{Answer Choices} \\
\midrule
\multicolumn{3}{l}{\textbf{Presence/Detection}} & \multicolumn{3}{l}{\textbf{Distribution/Segmentation}} \\
PD1 & Are active thermal hotspots detected? & Yes / No & DS1 & What is the spatial distribution of the active hotspots? & \makecell[l]{Scattered / Concentrated / \\ Linear / No active hotspots} \\
PD2 & Is smoke visible? & Yes / No & DS2 & How continuous is the fuel bed in the fire's potential path? & Continuous / Patchy / Discontinuous \\
PD3 & Are visible flames present? & Yes / No & DS3 & How consistent is the intensity of the active thermal hotspots in the scene? & \makecell[l]{Similar intensity / Different \\ intensity / No active hotspots} \\
PD4 & Are any buildings or residential structures visible? & Yes / No & DS4 & What proportion of visible vegetation is affected by fire, either actively burning or already burned? & 1--25\% / 25--50\% / $>$50\% / None \\
PD5 & Are natural fuel breaks like rock outcroppings or sparse vegetation visible? & Yes / No & DS5 & Approximately what proportion of the image is covered by above-ground vegetation such as shrubs and trees? & \makecell[l]{1--25\% / 25--50\% / 50--75\% / \\ 75--100\% / None} \\
PD6 & Are there standing dead trees that could contribute to high-intensity burning? & Yes / No & DS6 & What percentage of the RGB image is obstructed by smoke? & \makecell[l]{1--25\% / 25--50\% / 50--75\% / \\ 75--100\% / No smoke} \\
PD7 & Are there isolated heat sources far from the main fire perimeter? & Yes / No / No fire & DS7 & What percentage of the full scene exceeds 400 degrees Celsius? & $<$2\% / 2--4\% / 4--6\% / $>$6\% / None \\
PD8 & How many emergency vehicles are visible in the scene? & 0 / 1--2 / 3--4 / $>$4 & DS8 & What percentage of the full scene exceeds 200 degrees Celsius? & \makecell[l]{$<$5\% / 5--10\% / 10--15\% / \\ $>$15\% / None} \\
\midrule
\multicolumn{3}{l}{\textbf{Classification}} & \multicolumn{3}{l}{\textbf{Cross-Modal Reasoning}} \\
CL1 & What is the dominant fire behavior observed in the scene? & \makecell[l]{Active fire / Smoldering / \\ Extinguished / No fire} & CMR1 & What is the level of tree canopy obstruction of the fire's base? & \makecell[l]{Fully / Partially / Not \\ obstructed / No fire} \\
CL2 & What is the dominant vegetation type in the scene? & \makecell[l]{Coniferous / Deciduous / \\ Grassland / Shrubland} & CMR2 & What is the primary limitation to observing the active burn area in this scene? & \makecell[l]{Smoke / Canopy / Viewpoint / \\ No major limitations / No fire} \\
CL3 & Which moisture level best describes the live vegetation? & \makecell[l]{Lush/Green / Transitioning / \\ Dry/Cured} & CMR3 & What is the level of smoke obstruction of the fire's base? & \makecell[l]{Fully / Partially / Not \\ obstructed / No fire} \\
CL4 & What is the density of the forest canopy? & \makecell[l]{Dense/Closed / Moderate / \\ Sparse/Open / No forest} & CMR4 & What is the temperature of the hottest part of the fire in this scene in degrees Celsius? & \makecell[l]{100--200 / 200--300 / 300--400 / \\ 400--500 / $>$500 / No hotspots} \\
\cmidrule(r){4-6}
CL5 & What is the primary fuel type on the ground by overall coverage? & \makecell[l]{Grass / Forest litter / \\ Shrubs / Mixed} & \multicolumn{3}{l}{\textbf{Flight Planning}} \\
CL6 & How accessible is the active fire area via roads or trails? & \makecell[l]{Clear / Partially / No \\ clear access / No fire} & FP1 & What is the camera's viewing angle? & Nadir (top-down) / Oblique (angled) \\
\midrule
\multicolumn{3}{l}{\textbf{Localization/Direction}} & FP2 & What is the estimated flight altitude category? & 0--50 m / 50--100 m / 100--150 m / $>$150 m \\
LD1 & Where is the most intense hotspot located within the frame? & \makecell[l]{TL / TR / BL / BR / \\ Center / No hotspots} & FP3 & What is the current level of safety risk of the UAV's position near flames or smoke? & \makecell[l]{High risk / Medium risk / \\ Low risk / No fire} \\
LD2 & Where is the densest vegetation located? & \makecell[l]{TL / TR / BL / BR / \\ Center / Uniform / No veg} & FP4 & At the UAV's current flight altitude, which scene feature is the biggest risk to safe or consistent UAV movement? & \makecell[l]{A. Rugged terrain B. Uneven forest \\ C. Smoke columns D. No obstacles} \\
LD3 & From which region of the image does the largest smoke plume originate? & \makecell[l]{TL / TR / BL / BR / \\ Center / Spread / No smoke} & & & \\
LD4 & What is the primary location of the man-made structures? & \makecell[l]{TL / TR / BL / BR / \\ Center / No structures} & & & \\
\bottomrule
\end{tabularx}
\end{table*}

\section{Multimodal Inputs and Their Roles}
\label{sec:multimodal_inputs}

As described in the main paper, each WildFireVQA sample contains three aligned modalities: an RGB image, a color-mapped thermal visualization, and a radiometric thermal TIFF. The RGB image provides visible-spectrum scene context, while the color-mapped thermal visualization provides a human-readable view of thermal structure. The paired radiometric thermal TIFF further provides per-pixel temperature values, enabling temperature-grounded reasoning and stronger validation of fire-related queries.

These modalities play different roles during dataset construction and benchmark evaluation. During dataset construction, answer generation uses the RGB image, the aligned color-mapped thermal visualization, and a compact radiometric summary derived from the paired thermal TIFF. During benchmark evaluation, the main paper studies controlled settings that isolate the effects of input modality and retrieval augmentation. More specifically, models are evaluated under RGB, Thermal, and corresponding retrieval-augmented variants in which compact radiometric thermal statistics are appended as auxiliary context.

This distinction is important for interpreting the benchmark. The annotation pipeline uses richer multimodal context to improve label reliability, whereas the evaluation protocol is designed to analyze how different MLLMs use RGB appearance, thermal structure, and retrieved numerical thermal cues when answering wildfire monitoring questions.

\section{Annotation Protocol}
\label{sec:annotation_protocol}

\label{sec:annotation_prompt}

For dataset-wide answer generation, the prompt follows the structure described in the main paper. Each prompt contains the RGB image, the aligned color-mapped thermal visualization, the question text, the candidate answer options, and a compact radiometric summary derived from the paired thermal TIFF. The radiometric summary includes the minimum temperature, maximum temperature, temperature standard deviation, percentage of pixels exceeding $200^\circ$C, and percentage of pixels exceeding $400^\circ$C. These quantities provide compact, physically grounded thermal context in addition to the visual evidence. A simplified form of the annotation prompt is shown below.

\paragraph{Prompt Template}
\begin{quote}
You are provided with two aligned images of the same wildfire scene:

1. The first image is a standard RGB aerial image. \\
2. The second image is a color-mapped thermal image derived from radiometric thermal data.

Use both images together to understand wildfire activity in the scene.

You are also given a compact temperature summary computed from the paired radiometric thermal TIFF:

- Minimum temperature: \texttt{\{min\}} \\
- Maximum temperature: \texttt{\{max\}} \\
- Temperature standard deviation: \texttt{\{std\}} \\
- Percentage of pixels above 200$^\circ$C: \texttt{\{pct\_200\}} \\
- Percentage of pixels above 400$^\circ$C: \texttt{\{pct\_400\}}

Use the visual evidence and the temperature summary jointly, then answer the multiple-choice question by selecting the correct option.
\end{quote}

This prompt structure follows the dataset-wide answer generation procedure described in the main paper and preserves a consistent multimodal labeling protocol across all samples and question types.

\subsection{Example Annotation Prompt}
\label{sec:prompt_example}

\begin{figure*}[t]
\centering
\fbox{
\begin{minipage}{0.96\linewidth}

\textbf{Prompt:}

You are analyzing aerial wildfire imagery captured by a UAV.  
Two aligned images of the same scene are provided.

\begin{itemize}
\item The first image is a standard \textbf{RGB aerial image}.
\item The second image is a \textbf{radiometric thermal image rendered with an inferno colormap}.
\end{itemize}

Use \textbf{both images together} to understand wildfire activity in the scene.

\textbf{Temperature Summary (°C):}

\begin{itemize}
\item Minimum Temp: 32.1
\item Maximum Temp: 612.5
\item Mean Temp: 96.4
\item Top 3\% Mean: 428.2
\end{itemize}

Use these statistics as anchor points to relate thermal colors to approximate temperatures.

\vspace{0.3cm}

\begin{center}
\begin{tabular}{cc}
\includegraphics[width=0.35\linewidth]{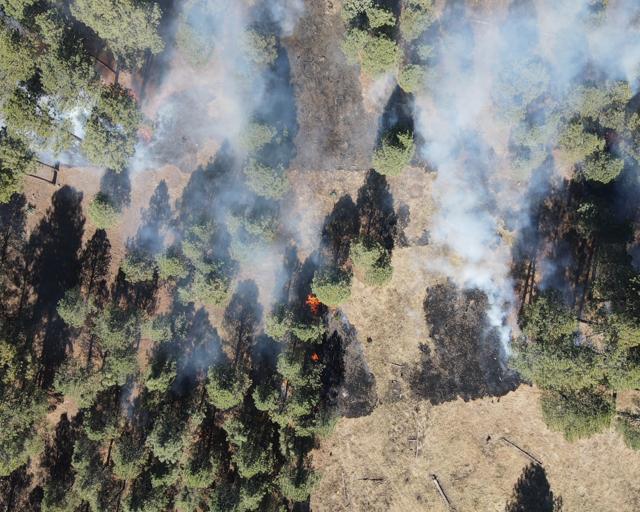} &
\includegraphics[width=0.35\linewidth]{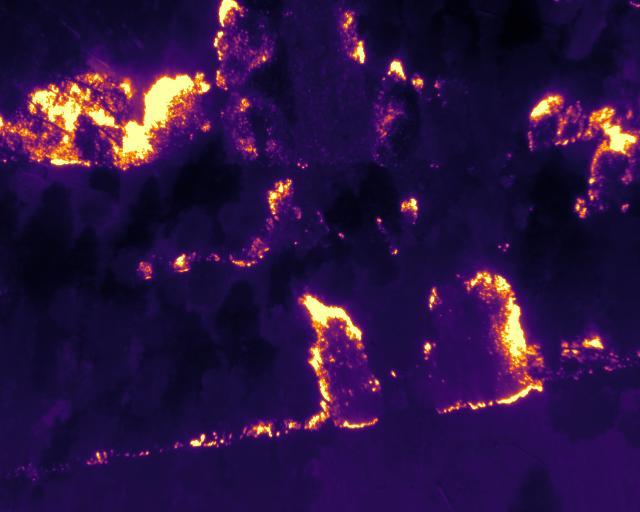} \\
\textbf{RGB Image} & \textbf{Thermal Image}
\end{tabular}
\end{center}

\vspace{0.3cm}

\textbf{Question:}

What is the spatial distribution of active hotspots in the scene?

\begin{itemize}
\item A. No active hotspots
\item B. Concentrated hotspots
\item C. Scattered hotspots
\item D. Linear fireline
\end{itemize}

\vspace{0.3cm}

\textbf{Example Model Responses}

\vspace{0.2cm}
{\color{green!60!black}
\textbf{Qwen3-VL-8B:}  
The thermal image reveals several spatially separated regions of elevated temperature distributed across the scene. These hotspots do not form a continuous line and are not concentrated in a single cluster.

\textbf{Answer: C. Scattered hotspots \ \ \checkmark}
}

\vspace{0.2cm}

{\color{red!70!black}
\textbf{LLaVA-v1.6-7B:}  
The RGB image shows flames concentrated in one region, and the thermal map appears dominated by a single hotspot.

\textbf{Answer: B. Concentrated hotspots \ \ \textbf{\xmark}}
}

\vspace{0.2cm}

{\color{red!70!black}
\textbf{InternVL2-8B:}  
The thermal pattern forms an elongated region of heat suggesting a linear fire structure.

\textbf{Answer: D. Linear fireline \ \ \textbf{\xmark}}
}

\vspace{0.2cm}

{\color{red!70!black}
\textbf{MiniCPM-V2:}  
The thermal visualization does not clearly indicate strong hotspots above background levels.

\textbf{Answer: A. No active hotspots \ \ \textbf{\xmark}}
}
\end{minipage}
}

\caption{
Example WildFireVQA prompt and model responses. The prompt contains aligned RGB and thermal images, a radiometric temperature summary derived from the thermal TIFF, and a multiple-choice question. Different multimodal language models may interpret the scene differently based on visual and thermal cues.
}
\label{fig:prompt_example}

\end{figure*}

Figure~\ref{fig:prompt_example} shows an illustrative prompt following the format used during WildFireVQA dataset construction. The prompt combines aligned RGB and thermal inputs with a compact radiometric summary derived from the paired thermal TIFF, then presents the question in multiple-choice form. This standardized structure is used to support consistent dataset-wide answer generation across the benchmark.

\end{document}